\documentclass[conference]{IEEEtran}
\usepackage{cite}
\usepackage{amsmath,amssymb,amsfonts}
\usepackage{algorithmic}
\usepackage{graphicx}
\usepackage{textcomp}
\usepackage{xcolor}
\usepackage{subfig}
\def\BibTeX{{\rm B\kern-.05em{\sc i\kern-.025em b}\kern-.08em
    T\kern-.1667em\lower.7ex\hbox{E}\kern-.125emX}}
\begin{document}

\title{Facial Expression Recognition using Facial Landmark Detection and Feature Extraction via Neural Networks}

\author{\IEEEauthorblockN{Fuzail Khan}
\IEEEauthorblockA{Department of Electronics \\ and Communication Engineering \\
National Institute of Technology Karnataka, Surathkal\\
Mangalore, India 575025 \\}}

\maketitle

\begin{abstract}
The proposed framework in this paper has the primary objective of classifying the facial expression shown by a person. These classifiable expressions can be any one of the six universal emotions along with the neutral emotion. After the initial facial localization is performed, facial landmark detection and feature extraction are applied where in the landmarks are determined to be the fiducial features: the eyebrows, eyes, nose and lips. This is primarily done using state-of-the-art facial landmark detection algorithms as well as traditional edge and corner point detection methods using Sobel filters and Shi Tomasi corner point detection methods respectively. This leads to generation of input feature vectors being formulated using Euclidean distances and trained into a Multi-Layer Perceptron (MLP) neural network in order to classify the expression being displayed. The results achieved have further dealt with higher uniformity in certain emotions and the inherently subjective nature of expression.
\end{abstract}


\section{Introduction}
The implementation can be broadly categorized into four stages: face location determination stage, facial landmark detection stage, feature extraction stage and emotion classification stage. An appropriate facial database was to be obtained which serves as our training and our testing data set, essentially consisting of humans displaying labelled emotions in the images [1]. The first stage deals with face detection algorithms that are to be implemented for the facial recognition, which mainly deals with image pre-processing and normalizing the images to eliminate redundant areas. In the second stage, it is stated that our facial fiducial landmarks: eyebrows, eyes, nose and mouth, are identified as the critical features for emotion detection and their feature points are hence extracted to recognize the corresponding emotion. These feature points are extracted primarily from state-of-the-art facial landmark detection algorithms along with traditional edge detection and corner point detection algorithms in Sobel horizontal edge detection and the Shi Tomasi corner point detection for the purpose. The input feature vectors were then calculated out of the facial feature extracted points obtained. In the third stage, these input feature vectors are given as input to the MLP neural network that is trained to then classify what emotion is being shown by the human.  

\section{Related work} 
Facial Expression Recognition (FER) systems have been implemented in a multitude of ways and approaches. The majority of these approaches have been based on facial features analysis while the others are based on linguistic, paralanguage and hybrid methods. 
Ghimire et al. [2] used the concept of position-based geometric features and angle of 52 facial landmark points. First, the angle and Euclidean distance between each pair of landmarks within a frame are calculated, and then successive subtraction between the same in the next frame of the video, using a SVM on the boosted feature vectors.
The appearance features are usually extracted from the global face region [3] or different face regions containing different types of information [4,5]. 
Happy et al. [3] utilized features of salient facial patches to detect facial expression. This was done after extracting facial landmark features and then using a PCA-LDA hybrid approach for dimensionality reduction and performance improvement. \\
For hybrid methods, some approaches [6] have combined geometric and appearance features to complement the positive outcomes of each other and in fact, achieve better results in certain cases.
In video sequences, many systems [2,7,8] are used to measure the geometrical displacement of facial landmarks between the current frame and previous frame as temporal features. 
Szwoch et al. [9] recognized facial expression and emotion based only on depth channel from the Microsoft Kinect sensor without using a camera. Local movements in the facial region are seen as features and facial expressions are determined using relations between particular emotions. Similarly, Sujono et al. [10] used the Kinect sensor to detect the face region and for face tracking based on the Active Appearance Model (AAM). Polikovsky et al. [11] presented facial micro-expression recognition in videos captured from 200 frames per second (fps) high speed camera. This method divides the face regions into certain localized regions, and then a 3D Gradients Orientation Histogram is generated from the motion in each local region. \\
Shen et al. [12] used infra-thermal videos by extracting horizontal and vertical temperature differences from different face sub regions. The Adaboost algorithm with the weak classifiers of k-Nearest Neighbor is used for Expression Recognition. Some researchers [9,10,12,13,14] have tried to recognize facial emotions using infrared images instead of images illuminated by visible light because the degree of dependence of visible light images on illumination is considerably higher. Zhao et al. [13] used near-infrared (NIR) video sequences and LBP-TOP (Local Binary Patterns-Three Orthogonal Planes) feature descriptors. This study uses component-based facial features to combine geometric and appearance information of face. Conventional FER systems in general use considerably lower processing power and memory when balanced with deep learning based approaches and are thus, still being researched for use in real time mobile systems because of their low computational power and high degree of reliability and precision. 

\section{Data Collection}

After a comprehensive search of databases that suited our purpose, the Karolinska Directed Emotional Faces (KDEF) dataset [1] was used to test our approach. It was developed at Karolinska Institutet, Department of Clinical Neuroscience, Section of Psychology, Stockholm, Sweden. It consists of a set of 4900 images of 70 subjects - 35 male and 35 female, each showing the 6 basic expressions and 1 neutral expression. Each expression being photographed twice from 5 different angles. The original image size is 562 x 762 pixels. A sample of images showing all 7 expressions is shown in Figure 1. \\
The data is partitioned as 90:10 for training data:testing data. This ratio was chosen as the objective is to train the neural network with maximum data along with the fact that 490 images are relatively enough to test the accuracy of the algorithm.  

\begin{figure}[h]
\centering
\subfloat[Fear][Fear]{
\includegraphics[width=2.5cm, height=2.5cm]{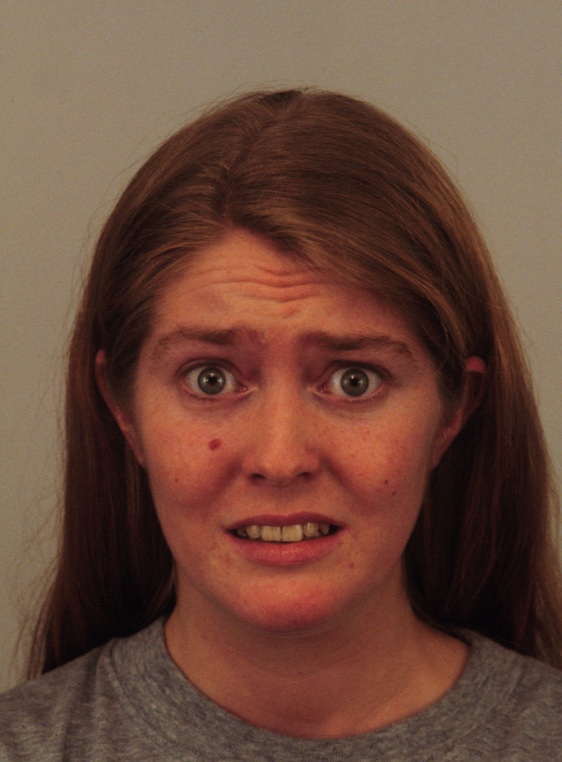}}
\subfloat[Anger][Anger]{
\includegraphics[width=2.5cm, height=2.5cm]{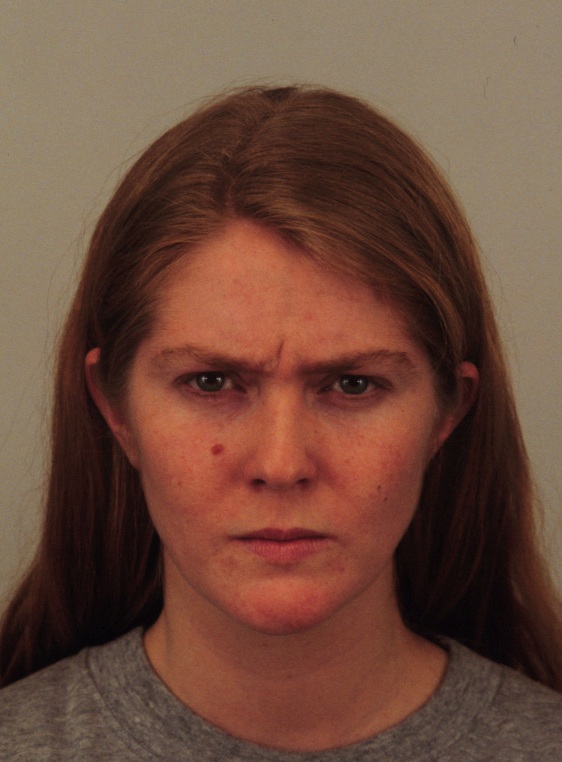}}
\subfloat[Disgust][Disgust]{
\includegraphics[width=2.5cm, height=2.5cm]{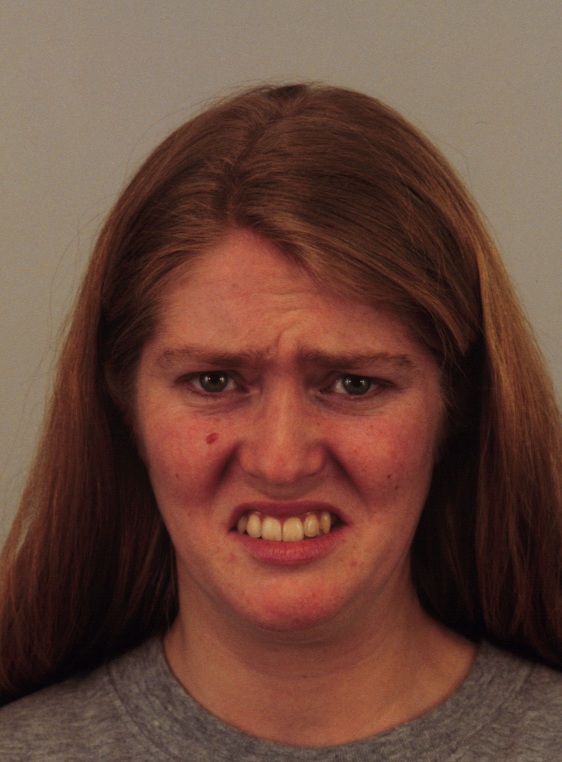}}
\qquad
\subfloat[Happiness][Happiness]{
\includegraphics[width=2.5cm, height=2.5cm]{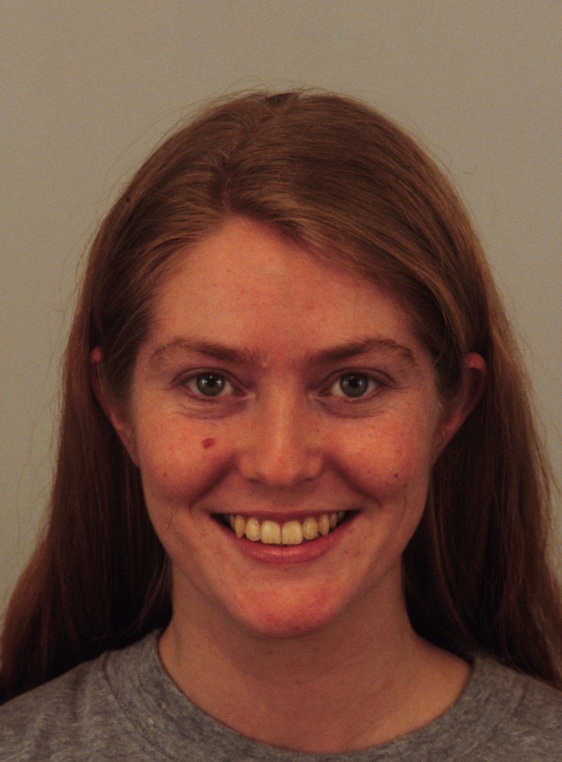}}
\subfloat[Sadness][Sadness]{
\includegraphics[width=2.5cm, height=2.5cm]{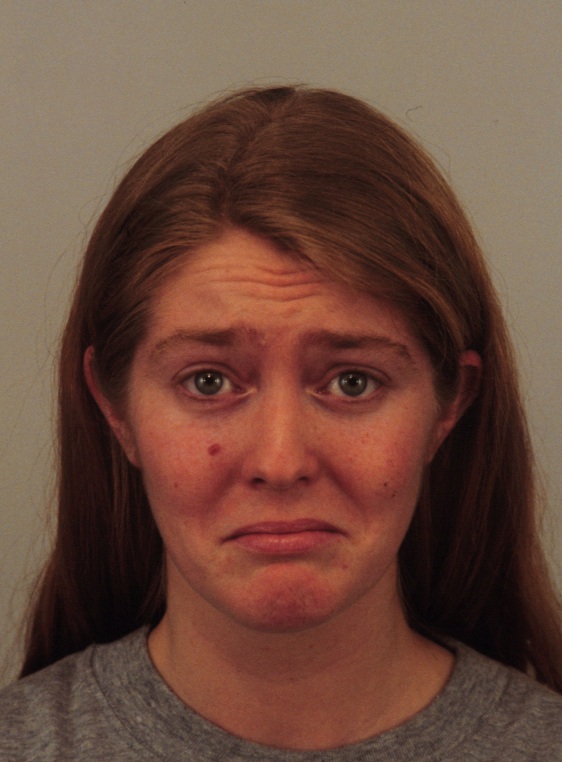}}
\subfloat[Surprise][Surprise]{
\includegraphics[width=2.5cm, height=2.5cm]{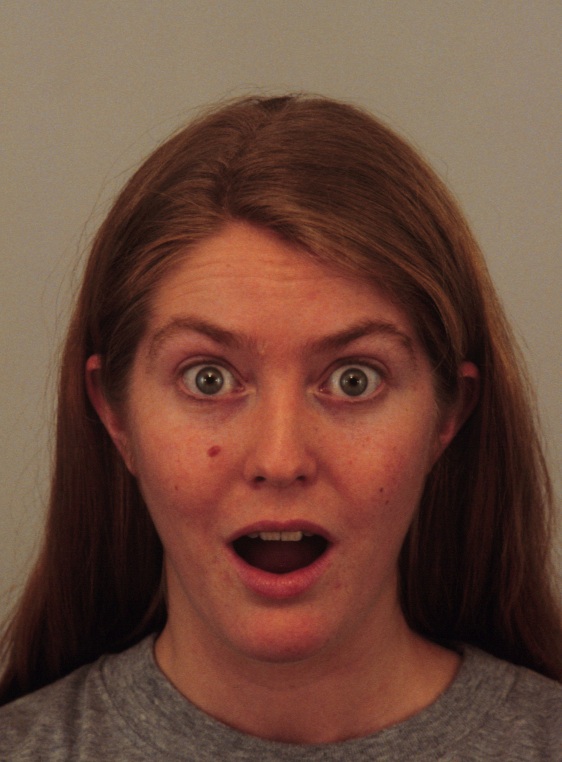}}
\qquad
\subfloat[Neutral][Neutral]{
\includegraphics[width=2.5cm, height=2.5cm]{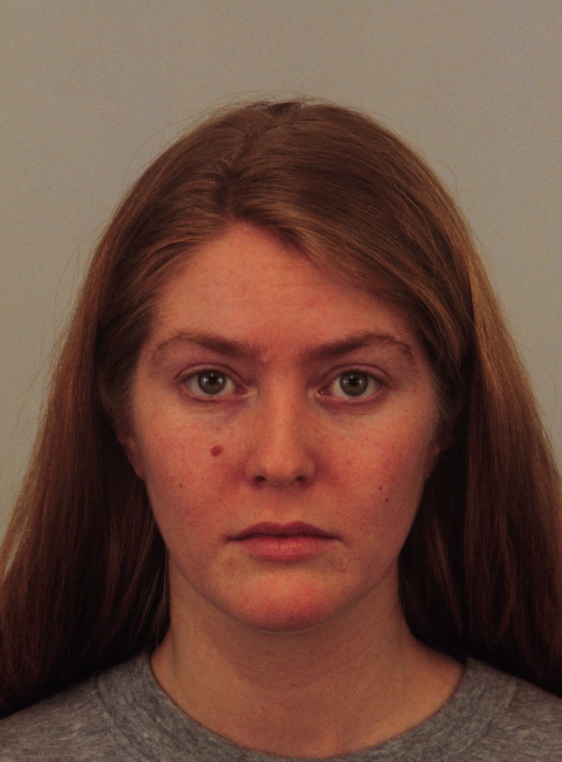}}
\qquad
\caption{Sample of images from the KFED facial database with 6 basic emotions and 1 neutral emotion being displayed.}
\label{fig:globfig}
\end{figure}

\begin{figure*}[h]
\centering
\includegraphics[width=18cm, height=4cm]{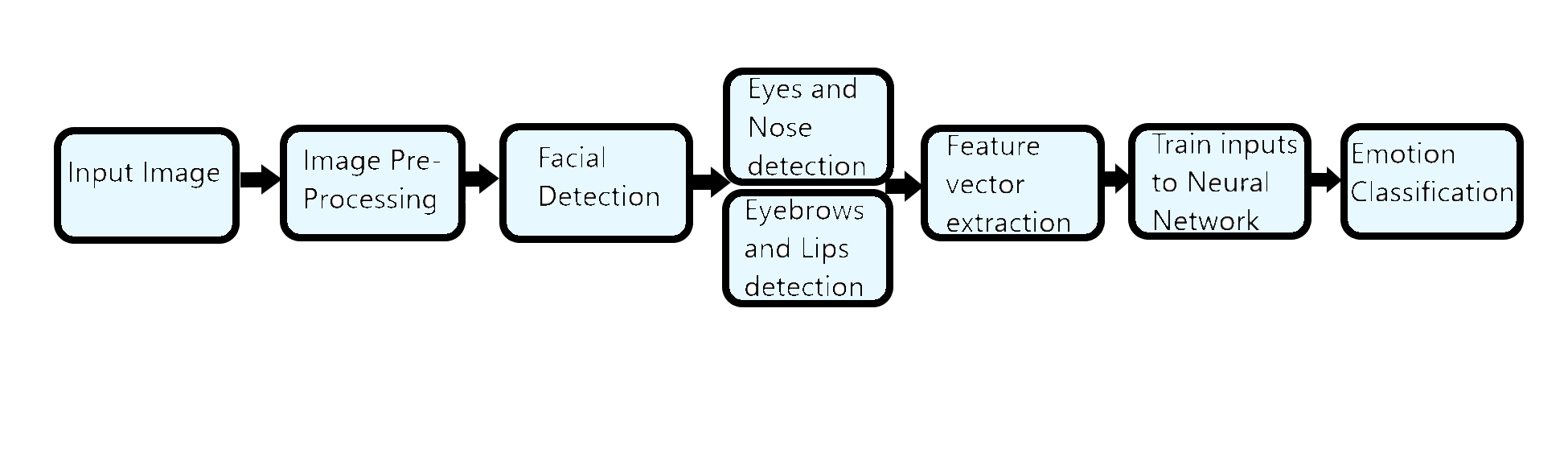}
\caption{Flowchart of the proposed FER methodology}
\end{figure*}

\section{Image pre-processing}

On getting the input image, the first step is to perform image pre-processing for the purpose of removing unwanted noise from the image and for enhancing the contrast of the image. A Low Pass 3x3 Gaussian filter was applied which helped smoothen the image and normalize gradient intensity values. Contrast Adaptive Histogram equalization was then carried out for illumination corrections. 
Normally, pre-processing would be to ensure uniform shape and size of the input images. This would not really apply to the KDEF database images as the facial orientation and location of the faces in each picture is uniform and hence, removes the need for any database specific pre-processing. 

\section{Facial Detection} 

In the facial detection stage, the objective is to find the face so as to limit our Region of Interest (RoI) such that all further processing takes place from that RoI onwards. 
This step of facial detection was accomplished using the now very reliable method of Haar classifiers. Haar feature-based cascade classifiers is an effective object detection method proposed by Paul Viola and Michael Jones [15]. \\
Each feature is a single value obtained by subtracting sum of pixels under white rectangle from sum of pixels under black rectangle. It determines features that best classifies the face and non-face images. Final classifier is a weighted sum of these weak classifiers as Adaboost combines many weak classifiers into one single strong classifier. \\
After the face is detected by this approach, the image is resized to only the face region and facial landmark detection takes place from this region onwards as seen in Fig. 4 b). 

\begin{figure}[h]
\centering
\subfloat[Raised brows]{
\includegraphics[width=2cm, height=2cm]{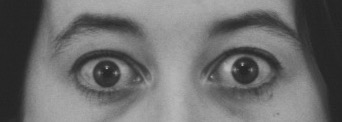}}
\subfloat[Inner brows tilt]{
\includegraphics[width=2cm, height=2cm]{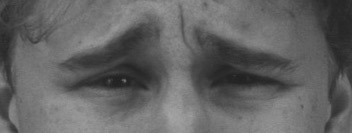}}
\subfloat[Nose wrinkle]{
\includegraphics[width=2cm, height=2cm]{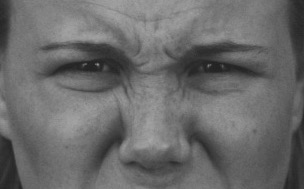}}
\subfloat[Tight lips]{
\includegraphics[width=2cm, height=2cm]{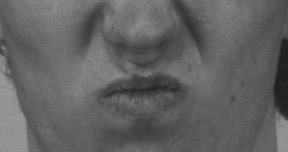}}
\newline 
\subfloat[Lips droop]{
\includegraphics[width=2cm, height=2cm]{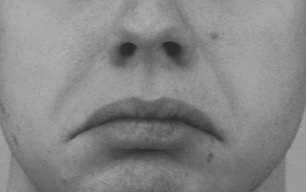}}
\subfloat[Mouth open]{
\includegraphics[width=2cm, height=2cm]{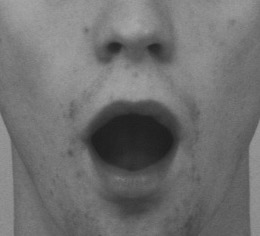}}
\subfloat[Mouth wide]{
\includegraphics[width=2cm, height=2cm]{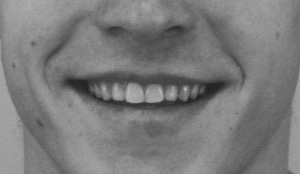}}
\qquad
\caption{Sample of images from the KFED facial database showing marked visible changes in facial landmarks for varied expressions.}
\label{fig:globfig}
\end{figure}

\section{Landmark detection}

\subsection{Eyes detection}
The purpose here is to get an Region of Interest (RoI) of the right eye and the left eye that is then to be used to extract feature points corresponding to the eyes.  The method decided was by using the circular Hough Transform acting on the input greyscale image of the face detected RoI. [16,17]
\\
The characteristic equation of a circle of radius r and center (a,b) is given by: 
	\begin{equation}
	    (x-a)^2 + (y-b)^2 = r^2
	\end{equation} 
Thus, the purpose of the Hough transform is to determine the triplet of parameters (a, b, r) which determines the final circles. To reduce this three-dimensional problem to a two-dimensional problem, the advantage of the circular Hough transform is used where it first detects only the center of the circle (a,b) with unknown radius on two-dimensional parameter space (x,y). The Hough transform thus generates the points in two-dimensional parameter space using image vector gradients, which is more efficient that applying the transform in ordinal three dimensional space.  \\ 
After this step, the radius of circle, r is determined for each local maximum in the (x, y) space. This is done on the basis of the strength of the gradient accumulated along each circle centered at the possible eye. \\
To further eliminate false detections arising out of the Hough transform, the eyes are selected such that the eye pair fulfills the below conditions: \\
(1) The true slant of the interpupillary line is within 20 degrees from the horizontal. \\
(2) The interpupillary distance is between 80 to 240 pixels.\\

As seen in Fig. 4 c), the steps are performed to localize the eye center or the pupil. A bounding box is then formed around the detected eyes taking the Hough Circle as the centroid of the box.
\subsection{Nose detection}
9 points for characterizing the nose are taken directly from the Intraface 49-point facial landmark detector and are verified in part by heuristic measures to ensure points are overlayed on the nose appropriately. 
 
\subsection{Eyebrows and lips detection} 

Intraface's 49 point landmark detector does an impressive job at localizing the points for the eyebrows (10 points) and lips (18 points). Although most facial landmark points can be directly taken from Intraface's output, there were a considerable cases where the detection wasn't satisfactory and so traditional edge detection methods were used to correct some landmark detections and give further confidence.
The eyebrows and upper lip always produce a distinct edge which can be detected using a edge detector. This edge detection would be accurately performed by these algorithms where the widely used ones include Sobel, Canny, Prewitt and Roberts. \\
After experimenting with the above techniques, the gradient-based Sobel edge detector gave the best results and was decided upon for eyebrow and lip detection. The Sobel operator is based on convolving the image with an integer-valued filter in the horizontal and vertical direction that give us two images which at each point contain the vertical and horizontal derivative approximations of the source image [18,19]. For our purpose, since the edge detection for lips and eyebrows is concerned with approximating the derivative along the X axis, the horizontal kernel is used and hence the Sobel horizontal operator. Once applied, the essential quantities: the strength and orientation of the resultant edges, are obtained. On applying the operator, multiple horizontal peaks are detected and a threshold quantity was used to try and obtain only the strong edges.   \\
After performing the edge detection, surplus edges were still present. A trial-and-error process had to be performed to check the effectiveness of binary thresholding methods. The basis of this effectiveness was how well the excess edges were refined out leading us to focus on the relevant feature points. The Otsu's thresholding turned out to be the best approach in this sense, trumping other forms of binary thresholding techniques. It's bi-modal histogram nature facilitates the purpose. Further, these false components having an area less than a specified threshold were removed. 
\newpage
\section{Feature corner point extraction} 
From the detection windows obtained for the eyebrows and lips, these RoI windows are now used to obtain the corner point features, the Shi Tomasi corner point detector is implemented. \\
The Shi Tomasi method [20] determines which windows produce very large variations in intensity when moved in both X and Y directions, thus computing the X and Y gradients. With each such window found, a score R is computed. \\
With each corresponding window, there is a window function = w(x,y). \\
M = w(x,y) x I 
where I = \[
\begin{bmatrix}
    \Sigma_{(x,y)} I{_x}^{2} & \Sigma_{(x,y)} I_{x}I_{y} \\
    
    \Sigma_{(x,y)} I_{y}I_{x} & \Sigma_{(x,y)} I{_y}^{2}    \\
\end{bmatrix}
\]
 where $I_x$ and $I_y$ are image derivatives in the x and y direction respectively. \\
The value R for each window: 
\begin{equation}
    R = min(\lambda_1, \lambda_2) \\ 
    \label{1}
\end{equation} 
where $\lambda_1$ and $\lambda_2$ are the eigen values of M. \\
After applying a threshold to R, important corners are selected and marked. Tuning parameters such as the minimum quality of image corners, minimum Euclidean distance between the corners and the window size for computing the derivative covariation matrix, the corner points for the fiducial landmarks are extracted. 

The selected feature corner points are : \\
For the eyebrows : 10 feature points - 5 left side and 5 right side. \\ 
For the eyes : 12 feature points - 6 left side and 6 right side. \\
For the nose : 9 feature points. \\
For the lips : 18 feature points. \\ \\
The final image showing a few points of the overall points is as shown in Fig 4. 

\begin{figure}[h]
\centering
\subfloat[Input face image][Input face image]{
\includegraphics[width=2.5cm, height=3cm]{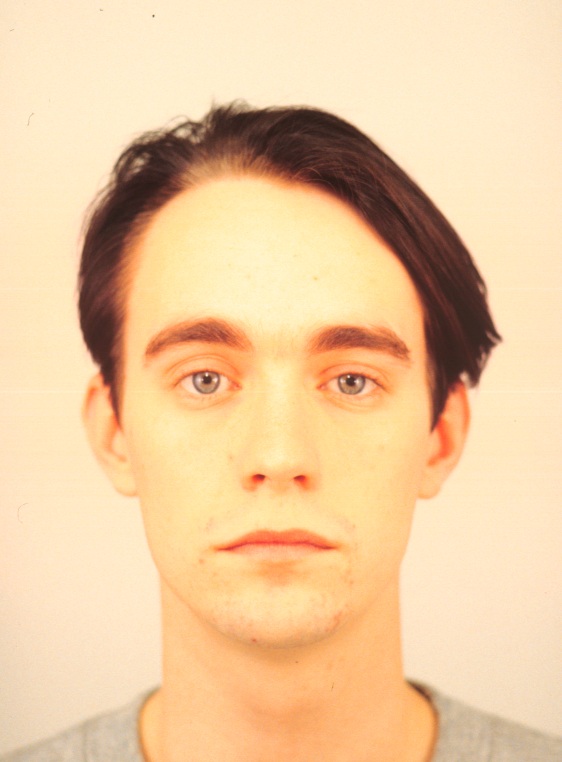}}
\vspace{1mm}
\subfloat[Face detected][Face detected]{
\includegraphics[width=2.5cm, height=3cm]{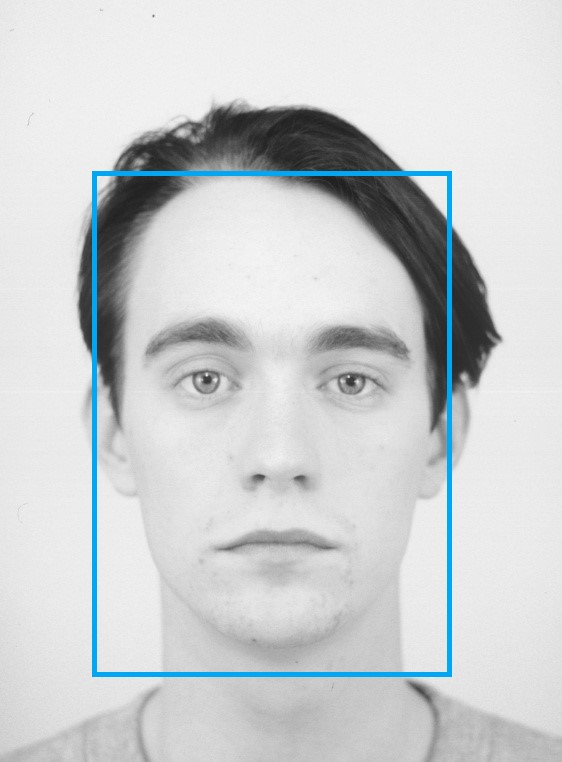}}
\vspace{1mm}
\subfloat[Eyes and nose RoI detected]{
\includegraphics[width=2.5cm, height=3cm]{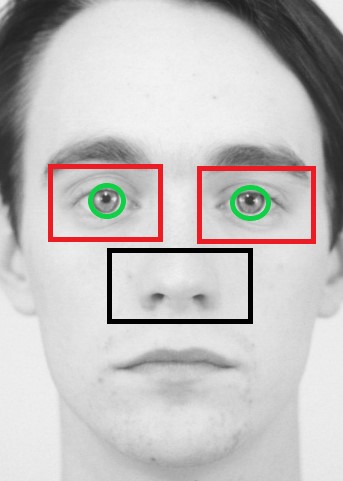}}
\qquad
\subfloat[Eyes and nose feature points detected]{
\includegraphics[width=2.5cm,height=3cm]{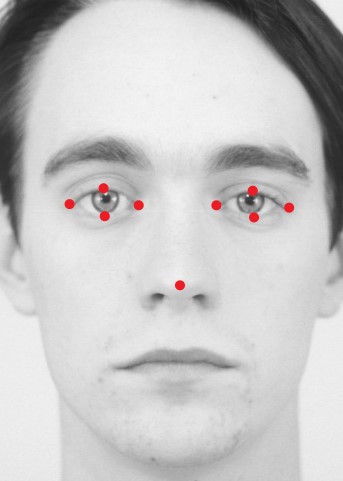}}
\vspace{1mm}
\subfloat[Eyebrow feature points detected][Eyebrow feature points detected]{
\includegraphics[width=2.5cm, height=3cm]{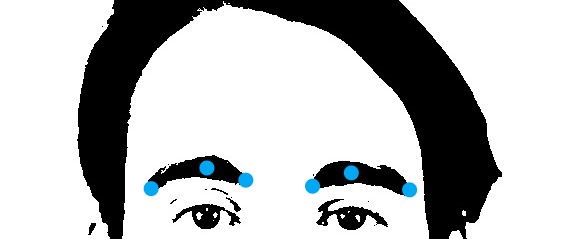}}
\vspace{1mm}
\subfloat[Lips feature points detected][Lips feature points detected]{
\includegraphics[width=2.5cm, height=3cm]{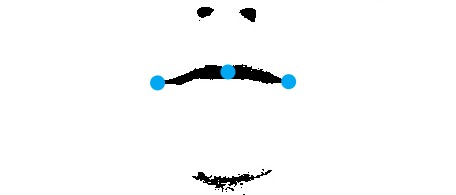}}
\qquad
\subfloat[Final image with a few feature points shown][Final image with a few feature points shown]{
\includegraphics[width=2.5cm, height=3cm]{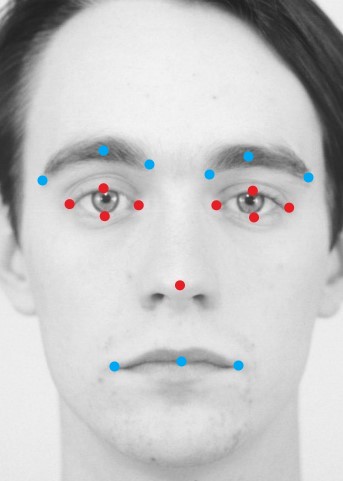}}
\caption{Various stages of facial landmark detection and feature extraction with few representative points being shown.}
\label{fig:globfig}
\end{figure}

\section{Finding feature vectors}

Now that the feature points are extracted from the input image, the next step is to decide the input feature vectors that need to be fed into and used to train our neural network. We feed in the intra-landmark euclidean distance vectors for the 49 landmark points detected. Majority of them would have not as much useful information but the MLP network shall learn that. It was also found these intra-landmark distance vectors need to be complimented by distance to a certain frame of reference especially on normalized faces. Thus, distance vectors are also taken to the average landmark point on the face. These two sets of feature vectors will be used as input to the network. On the basis of these input feature vectors, the neural network will hence learn from these and use these inputs to make classification decisions of the final output emotion. A subset of chosen feature vector inputs are as seen in Table 1.

\begin{table}[h]
\label{features}
\footnotesize
\caption{Subsample of Input feature vectors and description}
\begin{tabular}{|l|c|}
\hline
Definition & Formula \\
\hline
Left eye Height & V1 = F1-F2 \\ 
\hline 
Left eye Width & V2 = F4 - F3 \\
\hline
Right eye Height & V3 = F5 - F6 \\
\hline
Right eye Width & V4 = F8- F7 \\
\hline
Left eyebrow width & V5 = F11 - F10 \\
\hline
Right eyebrow width & V6 = F14 - F13 \\
\hline
Lip width & V7 = F17 - F16 \\ 
\hline
Left eye upper corner and left eyebrow center dist. & V8 = F12 - F1 \\
\hline
Right eye upper corner and right eyebrow center dist. & V9 = F15 - F5 \\
\hline
Nose centre and lips centre dist. & V10 = F9 - F18 \\
\hline
Left eye lower corner and lips left corner dist. & V11 = F2 - F16 \\
\hline
Right eye lower corner and lips right corner dist. & V12 = F6 - F17 \\
\hline
\end{tabular}
\end{table} 

\section{Training the neural Network}

\subsection{Neural networks and back propagation} 
Neural networks have proved their ability in the recent past to deliver simple and powerful solutions in areas relating to signal processing, artificial intelligence and computer vision. The back-propagation neural network is a widely used neural network algorithm due to its simplicity, together with its universal approximation capacity. \\  
The back-propagation algorithm defines a systematic way to update the synaptic weights of multi-layer perceptron networks. The supervised learning is based on the gradient descent method, minimizing the global error on the output layer. The learning algorithm is performed in two stages: feed-forward and feed- backward. In the first phase the inputs are propagated through the layers of processing elements, generating an output pattern in response to the input pattern presented. In the second phase, the errors calculated in the output layer are then back propagated to the hidden layers where the synaptic weights are updated to reduce the error. This learning process is repeated until the output error value, for all patterns in the training set, are below a specified value. 
\subsection{The neural network configuration} 
The light-weight MLP neural network consists of an input layer, 2 hidden layers and an output layer. The first hidden layer has 100 neurons and the second hidden layer has 500, with a softmax output layer of 7 labels. \\
Regarding the activation functions for the nodes, non-linear activation functions are used for the neurons in an MLP network. To decide the activation function, it is necessary to examine the purpose and the output of the neural network: \\ \\
P(Y$_i$/x) = Probability of Emotion (Y$_i$) given the input image x, where i $\epsilon$ [0,6] corresponding to the 7 output emotions.\\ 
Sigmoid function :
\begin{equation}
    \phi(z) = \frac{1}{(1 + e^{-z})}
\end{equation}
For a multi-class classification, it is suitable to use the softmax Sigmoid activation function where a k-dimensional vector is the output for a given k-class classification where each of the k values $\epsilon (0,1)$ and sum up to 1 : 
\begin{equation}
    \sigma : R^k \longrightarrow 
    \Bigg\{ 
    \sigma \hspace{1mm} \epsilon \hspace{1mm} R^k \hspace{1mm} |\hspace{1mm} \sigma_i > 0\hspace{1mm};  \hspace{5mm} \Sigma_{i=1}^{K} = 1 
\Bigg\}
\end{equation}

\begin{equation}
    \sigma(z)_j = \frac{e^{z^j}}{\Sigma_{k=i}^{K} e^{z^j}} \hspace{1mm} ; \hspace{5mm}
    j \epsilon [1,k]
\end{equation} \\ \\
Thus, the final emotion classified =\\ max(P(Y$_i$/x)) $\forall$ i $\epsilon$ [0,6] i.e the maximum of the probabilities of all the 7 emotions given the image x. \\ \\  
After the process of parameter tuning, optimization and regularization, the neural network was configured by the optimal learning rate = 0.005 with Adam optimizer (0.9,0.999) and Dropout probability of 0.3.

\newpage
\section{Results and Conclusion}

After training the neural network as explained above, our testing set of images from the KDEF database was used to check the performance of the proposed FER system. The results of the test have been presented as a confusion matrix as shown in Table 2 and the false positive detection rates per emotion as shown in Table 3. 

\begin{table}[h]
\label{tab:my_label}
\centering
\caption{Confusion matrix of emotion classification}
\fontsize{6}{8}\selectfont
\begin{tabular}{|c|c|c|c|c|c|c|c|}
\hline
 I/O & Happiness & Anger & Disgust & Surprise & Fear & Sadness & Neutral  \\
\hline
Happiness & 98.2\% & 0\% & 0\%  & 0\% & 1.5\% & 0\% & 0.3\% \\
\hline
Anger & 0\% & 85.6\% & 9.3\% & 2.6\%  & 1.7\%  & 0.8\%  & 0\% \\
\hline
Disgust & 0\% & 4.7\% & 84.9\%  & 1.1\%  & 1.1\%  & 7.0\%  & 1.2\% \\ 
\hline
Surprise & 0\% & 0.4\% & 1.1\%  & 95.8\%  & 1.3\%  & 1.4\%  & 0\% \\
\hline
Fear & 1.2\% & 1.3\% & 4.2\%  & 1.7\%  & 86.5\%  & 2.4\%  & 2.7\% \\
\hline
Sadness & 0\% & 1.1\% & 0.4\% & 0.4\%  & 9.6\% & 86.6\%  & 1.9\% \\
\hline
Neutral & 0.6\% & 0.5\% & 2.1\%  & 0\%  & 2.4\%  & 4.3\%  & 90.1\% \\ 
\hline
\end{tabular}
\end{table} 

From Table 2 and Table 3, it is observed that the emotions of happiness and surprise are being detected relatively well with 95\%+ true positive success rates, which indicates that the facial reactions to when a person is happy and surprised are more uniform than the other emotions leading to high success rates. The emotions of anger, sadness, disgust and fear have their success rates in the 84-86\% range and have an overlap among almost all of the other emotions indicating that people may have the same facial reaction for two different emotions as well as having different facial reactions for the same emotions, which leads to the overlap among these classified emotions, while the neutral emotion is being recognized quite well with a 90.1\% success rate. \\

\begin{table}[h]
\label{tab:my_label}
\begin{center}
\centering
\caption{False positive rate per emotion}
\begin{tabular}{|c|c|}
\hline
Emotion & False positive rate \\
\hline
Happiness & 1.8\% \\
\hline
Anger & 14.4\% \\
\hline
Disgust & 15.1\% \\ 
\hline
Surprise & 4.2\% \\ 
\hline
Fear & 13.5\% \\ 
\hline
Sadness & 13.4\% \\ 
\hline
Neutral & 9.9\% \\ 
\hline
\end{tabular}
\end{center}
\end{table} 

It is well known that expression is inherently subjective and different people react in varied manners to different emotions. For example, person A looking afraid might look similar to person B looking disgusted. Our conclusions of happiness and surprise having lesser false positive detection rates can be used to further analyse this concept. As a preliminary act of just scratching the surface, the focus is on feature vectors associated with the lips. As noted in Table 3, these relevant feature vectors would be V7, V10, V11 and V12. These feature vectors are essentially the euclidean distance between two extracted feature points. These can be used to differentiate between different subjects showing varied degrees of expression. With the intention of maintaining a 50-50 split, the median shall be used to split the data on these feature vectors. This concept applies with the implicit assumption that, for instance, the wider the smile, hence the greater the feature vector and the more expressive the person. Fig. 5 illustrates our notion of subjective expressions in the data. 

\begin{figure}[h]
\centering
\subfloat[A smile to a greater degree]
{\includegraphics[width=3cm, height=3cm]{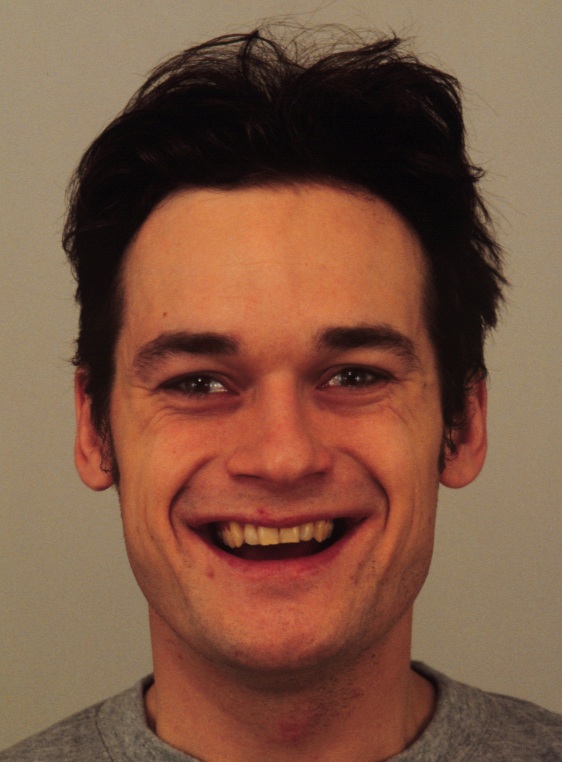}}
\hspace{0.5cm}
\subfloat[A smile to a lesser \mbox{degree}]{
\includegraphics[width=3cm, height=3cm]{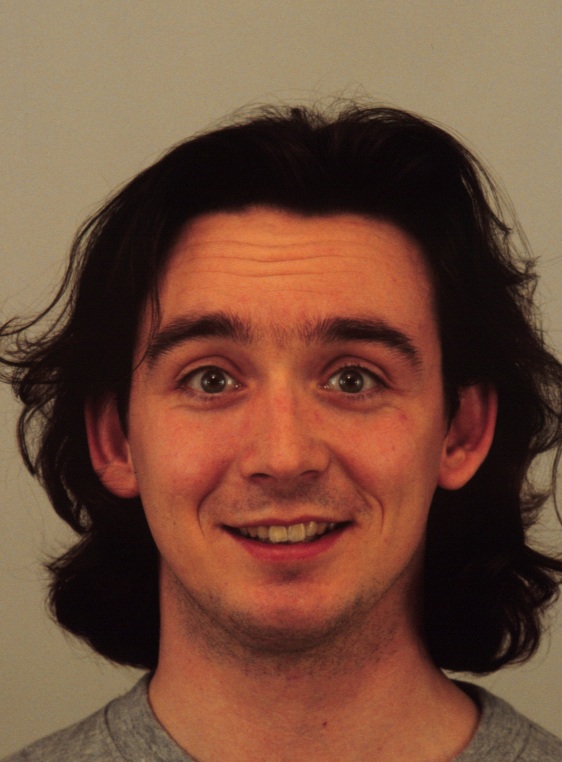}}
\caption{The emotion of happiness being conveyed with different extents of expression.}
\label{fig:globfig}
\end{figure}

Experimenting with these two sets of data showed better results among the two groups, especially with the emotions of anger and fear. As was expected, the 'more expressive' people display a higher degree of correlation in expressing emotions among themselves and the same applies to the 'less expressive' group. Further exploration into the individuality and personal nature of facial expressions can better recognize emotions going forward. \\
On a larger note, a major prospective application would be to maximise use of FER systems in personal assistants such as Siri and Alexa. Since these personal assistants will be constantly learning solely on their respective user and since that user has certain mannerisms and behaviours unique to himself, the ability to correctly identify his or her moods and emotions will be significantly improved. \\
Considering future improvements, more data would lead to better results and so, the number of images in the facial expression databases is hence, a limiting factor. Also, the incidence of facial hair, occlusions and the like have not been taken into account and so, further analysis and studies are required to better performance by dealing with these subjects.

\end{document}